\documentclass{article}
\usepackage{ijcai21}

\usepackage{times}
\usepackage{soul}
\usepackage{url}
\usepackage[hidelinks]{hyperref}
\usepackage[utf8]{inputenc}
\usepackage[small]{caption}
\usepackage{graphicx}
\usepackage{amsmath}
\usepackage{amsthm}
\usepackage{booktabs}
\usepackage{algorithm}
\usepackage{algorithmic}
\usepackage{subfigure}

\usepackage{bm}
\title{HIP Network: Historical Information Passing Network for Extrapolation Reasoning on Temporal Knowledge Graph}

\author{
Yongquan He$^{1,2}$
\and
Peng Zhang$^{1}$\footnote{Corresponding Author}\and
Luchen Liu$^{1,2}$\and
Qi Liang$^1$\and
Wenyuan Zhang$^{1,2}$\And
Chuang Zhang$^1$
\affiliations
$^1$Institute of Information Engineering, Chinese Academy of Sciences\\
$^2$School of Cyber Security, University of Chinese Academy of Sciences\\
\emails
\{heyongquan, pengzhang, liuluchen, liangqi, zhangwenyuan, zhangchuang\}@iie.ac.cn
}

\begin{document}

\maketitle

\begin{abstract}
  In recent years, temporal knowledge graph (TKG) reasoning has received significant attention.
  Most existing methods assume that all timestamps and corresponding graphs are available during training, which makes it difficult to predict future events.
  To address this issue, recent works learn to infer future events based on historical information.
  However, these methods do not comprehensively consider the latent patterns behind temporal changes, to pass historical information selectively, update representations appropriately and predict events accurately.
  In this paper, we propose the \textbf{H}istorical \textbf{I}nformation \textbf{P}assing (HIP) network
  to predict future events.
  HIP network passes information from temporal, structural and repetitive perspectives, which are used to model the temporal evolution of events, the interactions of events at the same time step, and the known events respectively.
  In particular, our method considers the updating of relation representations and adopts three scoring functions corresponding to the above dimensions.
  Experimental results on five benchmark datasets show the superiority of HIP network, and the significant improvements on Hits@1 prove that our method can more accurately predict what is going to happen.
\end{abstract}

\section{Introduction}
Knowledge graphs (KGs) are multi-relational graphs,
where nodes and various types of edges reflect entities and relations respectively. 
Each edge is presented as a triple of the form $(subject, relation, object)$, e.g., $(Obama, visit, China)$.
Due to most of the KGs are far from complete,
knowledge graph reasoning is proposed to infer missing facts \cite{survey3}.
And this problem has been extensively studied for static KGs.
The common way is to embed entities and relations into the continuous vector spaces, 
and computing a score for each triple by applying a scoring function to these embeddings \cite{TransE,DistMult,CompIEX}.
However, facts may not always be true in the real world,
which introduces the concept of the temporal knowledge graph (TKG).
And each fact can be seen as a quadruple which takes the timestamp into consideration, i.e., $(subject, relation, object, timestamp)$.
Reasoning on TKGs is more complex because of its native dynamic nature.

To solve this problem, previous works attempt to extend the static KG embedding methods which score the likelihood of missing facts with timestamps \cite{TTransE,HyTE,TA-DistMult,DE-SimpIE}.
But these methods neglect the structure during learning the representations.
So methods like TeMP \cite{TeMP} are proposed recently.
They focus on the evolving graph snapshots at multiple time steps for the dynamic representation and inference.
However, they are not suitable for predicting future events, as the timestamps and the corresponding graphs are unknown.
This problem is called \textbf{extrapolation reasoning}, which aims to predict new facts over future timestamps \cite{RENET}.
Extrapolation reasoning over TKGs is less studied but significant,
since forecasting emerging events is useful for real-world applications.

\begin{figure}
    \centering
    \includegraphics[width=2.6in]{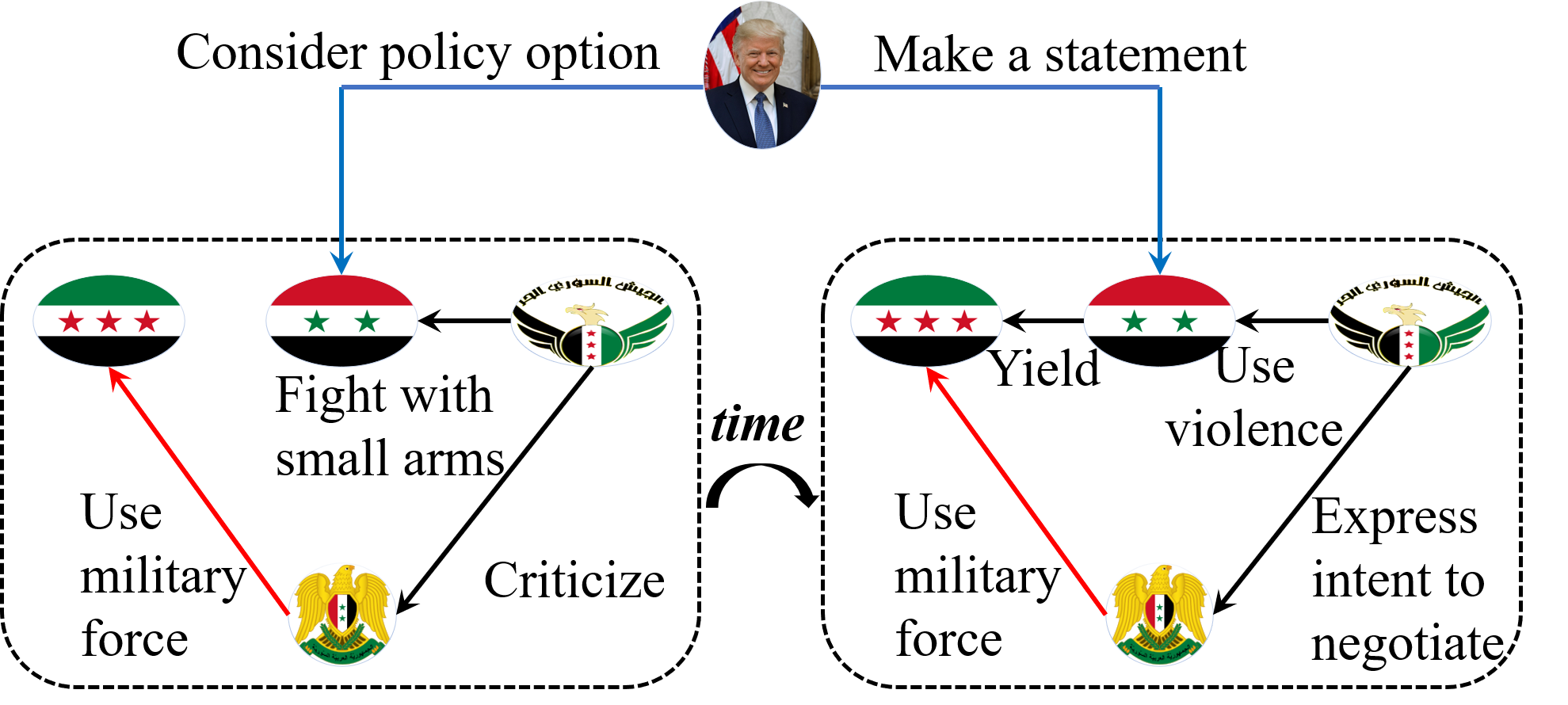}
    \caption{Temporal knowledge subgraphs about the evolution of political events in Syria and the Trump government's responses.
}
    \label{fig:intro}
\end{figure}

\begin{figure*}
    \centering
    \includegraphics[width=\linewidth]{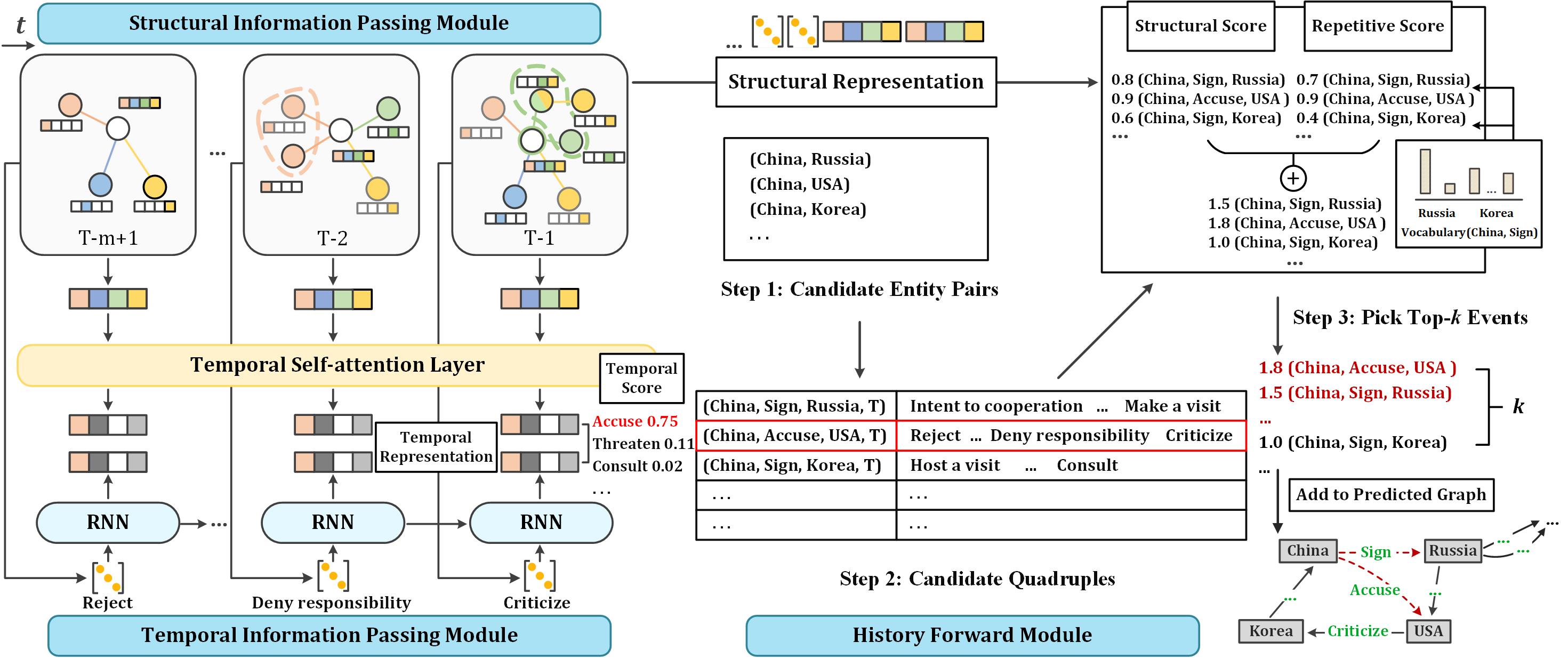}
    \caption{The framework of our HIP network. HIP network contains three components. The structural information passing module to capture neighborhood interactions. The temporal information passing module to model evolution patterns of events. And the HIP module to calculate the plausibility of
events from the repetitive, structural and temporal perspectives.
}
    \label{fig:framework}
\end{figure*}

Recently, RE-NET \cite{RENET} and CyGNet \cite{CyGNet} are proposed to tackle the extrapolation reasoning problem.
RE-NET uses RNNs \cite{RNN} and RGCN-based aggregator \cite{RGCN} to encode historical information for future predictions.
And CyGNet proposes a copy-generation mechanism, which generates future events based on the historical distribution.
However, there are still several challenges not fully addressed.
As the patterns associated with the occurrence of events are sophisticated, how to disentangle and model these patterns is useful and challenging.
As shown in Figure~\ref{fig:intro}, we can see the temporal evolution of events (blue edges), as policy options are often considered before making a statement.
And at the same time step, the co-occurring events in a neighborhood structure also have dependencies and interactions (dotted lines).
Moreover, events may occur repeatedly along the history (red edges).
These phenomenons highlight the importance of learning different patterns to access more reference information.
Besides, due to the dynamic nature of TKGs, only a part of entities are active at each time step.
And the relations between entity pairs also change.
Previous methods usually update the embeddings as static KG methods or only concentrate on the representations of entities,
which can not fully learn the time-sensitive features for both entities and relations.

In this paper, we propose \textbf{H}istorical \textbf{I}nformation \textbf{P}assing (HIP) network, a novel framework for extrapolation reasoning on TKGs.
In the structural information passing part, our method uses CompGCN \cite{CompGCN} to update the structural representations in a disentangled manner, which utilizes the relation embeddings and aggregates neighborhood information selectively.
As for the temporal part, we generate the relation embeddings for specific entity pairs to capture the temporal evolution,
and get the current embeddings for entities with the temporal self-attention mechanism.
We also select object entities from history, to make reference to repetitive facts.
Then HIP network adopts a multi-step reasoning algorithm with three score functions, to generate predictions in a sequential manner.

Our contributions are summarized as follows: 
1)We propose a novel learning framework to handle the extrapolation reasoning problem on TKGs, which passes historical information from the temporal, structural and repetitive perspectives.
2)Our method specifically considers the updating of relation embeddings in the information passing process and proposes a multi-step reasoning algorithm with three scoring functions, which can infer future events sequentially.
3) The experimental results on five TKG datasets achieve state-of-the-art performance, and the significant improvements on Hits@1 prove the effectiveness of our method.

\section{Related Work}

\paragraph{Static KG embeddings.}
There are increasing interests in knowledge graph embedding (KGE) methods,
which aim to embed entities (nodes) and relations (edges) into the continuous vector spaces,
such as the translating models \cite{TransE,TransE1} and the semantic matching models \cite{DistMult,CompIEX,RotatE}.
And some other models score facts based on the deep neural network with feed-forward or convolutional layers \cite{ConvE,RGCN,visual,CompGCN}.
There are also works that use external information, such as typing information \cite{typing} and logic rules \cite{rule1}.
More details can be found in recent surveys \cite{survey3}.
However, these methods are not able to predict future events, as all entities and relations are treated as static.

\paragraph{Temporal KG reasoning.}
Recent works attempt to extend the static KG embeddings methods with temporal information. 
HyTE \cite{HyTE} projects the entities and the relations onto timestamp specific hyperplanes.
DE-SimpIE \cite{DE-SimpIE} defines a function that takes an entity and a timestamp as input,
to generate time-specific representations.
And there are many other methods consider the timestamp to do the temporal reasoning \cite{TTransE,TA-DistMult}.
Another line of works try to capture changes in neighborhood structure,
which use temporal recurrence and graph neural network together \cite{TeMP,DySAT}.
However, the above methods are not suitable for extrapolation reasoning on TKGs, because the structure and timestamp of the future graph are unknown.
To solve this problem, RE-NET \cite{RENET} and CyGNet \cite{CyGNet} are proposed, which are the most relevant methods to our work.
RE-NET defines the joint probability distribution of all events in an autoregressive fashion.
And CyGNet proposes a copy-generation mechanism, which makes predictions with the historical vocabulary.

\section{Method}

This section introduces the proposed model, named HIP network, for extrapolation reasoning on TKGs.
We first define the task and notations, and give an overview of the model architecture.
Then we detail the individual components in the following sections.

\subsection{Task Definition and Model Architecture}

\paragraph{Notations and task definition.} A temporal knowledge graph (TKG) is composed of a sequence of time-stamped subgraphs,
i.e., $\mathcal{G}$ = $\{\mathcal{G}^{(1)},  \mathcal{G}^{(2)}, ...,  \mathcal{G}^{(T)}\}$, 
where $ \mathcal{G}^{(t)}$ = $\{\mathcal{E},\mathcal{R},\mathcal{O}^{t}\}$.
Here, $\mathcal{E}$ and $\mathcal{R}$ are known entities and relations across all time steps.
And $\mathcal{O}^t$ is the events (edges) set at time $t$, each event is denoted as $(s,r,o,t) $,
where $s,o \in \mathcal{E}$ and $r \in \mathcal{R}$.
Our task is to predict missing events about an object query $(s, r, ?, t+\Delta t)$ (or a subject query $(?, r, o, t+\Delta t)$), 
through using the historical information in $\{\mathcal{G}^{(t')}\} (t' \le t)$, 
even if the events in time period $\Delta t$ are unknown.

 \paragraph{Model architecture.} As shown in Figure~\ref{fig:framework}, our model has three main components,
namely structural information passing (SIP) module, temporal information passing (TIP) module,
and history forward (HF) module.
In the SIP module, we consider the interactions of co-occurring events in the latest graph through the CompGCN-based aggregator.
We divide the representations into multiple independent components, and update structural embeddings for the entities and relations according to the attention weight of each component.
And in the TIP module, we focus on capturing the temporal evolution patterns of the events between specific entity pairs and integrating structural embeddings across time.
We selectively incorporate historical representations for entities with a temporal self-attention layer, and use RNN to model evolution process of events to generate the temporal embeddings for relations.
Finally, in the HF module, we need to predict what events are likely to occur.
We adopt three scoring functions to evaluate the events from different perspectives.
Then we generate the predicted graph according to the reasoning results and move to the next time step.

\subsection{Structural Information Passing Module}

To capture the neighborhood interactions,
we update the structural embeddings of entities and relations based on the recently historical graph $ \mathcal{G}^{(t-1)}$.
However, due to the streaming nature of the temporal knowledge graph, when new neighbors come or disappear,
we need to update the related parts of embeddings while retaining useful information.
So our method uses CompGCN \cite{CompGCN} in a disentangled manner to selectively pass structural information.

In this module, we aim to learn a disentangled embedding $\bm{x}_{s,t}$ for entity $s$.
Firstly, we project $\bm{x}_{s,t-1}$ onto $K$ spaces as follows:
\begin{equation}
    \bm{h}_{s,k} = \bm{U}_{k}^{T}\bm{x}_{s,t-1},
\end{equation}
where $\bm{U}_{k} \in {\rm R}^{d_{in} \times \frac{d_{in}}{K}}$ is the parameter of channel $k$ and $K$ is the hyperparameter.
Then we can obtain the node embeddings of $K$ different components, i.e., $\bm{h}_s = \{ \bm{h}_{s,1}, \bm{h}_{s,2}, ..., \bm{h}_{s,k}, ..., \bm{h}_{s,K}\}$.

Unlike most graph neural networks which embed only nodes in the graph, CompGCN uses embeddings of relations instead of matrices \cite{RGCN,CompGCN}.
So for each quadruple $(s,r,o,t-1) \in \mathcal{G}^{(t-1)}$, we also need to project the results of entity-relation composition operation onto the $K$ spaces as follows:
\begin{equation}
    \bm{c}_{o,k} = \bm{V}_{k}^{T}\phi(\bm{x}_{r,t-1},\bm{x}_{o,t-1}),
\end{equation}
where $\bm{V}_{k} \in {\rm R}^{d_{in} \times \frac{d_{in}}{K}}$ is the parameter of channel $k$.
And $\phi(\bm{x}_{r},\bm{x}_{o})$ is the composition operation, such as subtraction, multiplication and circular-correlation.
Then we can set $K$ attention values, and the $k$-th attention value $\alpha_k$ indicates how related the message $\phi(\bm{x}_{r},\bm{x}_{o})$ is to the $k$-th component $\bm{h}_{s,k}$.
The attention weight $\alpha_k$ is computed as:
\begin{equation}
    \alpha_k = \frac{{\rm exp}({\rm ReLU}(\bm{W}[\bm{c}_{o,k}; \bm{h}_{s,k}]))}   {\sum_{k'=1}^{K} {\rm exp}({\rm ReLU}(\bm{W}[\bm{c}_{o,k'}; \bm{h}_{s,k'}]))},
\end{equation}
where $\bm{W} \in {\rm R}^{1 \times \frac{2d_{in}}{K}}$ is a transformation matrix that can be trained.
Then the update equation of the structural aggregator is given as:
\begin{equation}
    \bm{x}_{s,k}^{l} = \sum_{(r, o') \in \mathcal{N}(s)} \alpha_k  \bm{W}_{\lambda (r)}^{l,k}  \bm{c}_{o',k}^{l-1}.
\end{equation}
Here,  $\bm{W}_{\lambda (r)}^{l,k} \in {\rm R}^{\frac{out}{K} \times \frac{in}{K}}$ is a relation-type parameter about the $k$-th component for the original, inverse and self-loop relations.
Finally, We tile the $K$ output embeddings $\bm{x}_{s,k}^{L}$ to obtain the structural representation $\bm{x}_{s, t}$ of entity $s$ at time $t$, where $L$ is the last layer.
And as the CompGCN can consider the embeddings of relations in the aggregating process, we can also get the refined embedding $\bm{x}_{r,t}$ for relation $r$. 

\subsection{Temporal Information Passing Module}

This module seeks to model the events evolution patterns between entity pairs and integrate information across time, to generate the temporal embeddings for entities and relations.

We assume that the embeddings from the SIP module sufficiently capture local structural information at each time step, which enables a disentangled modeling of structural and temporal information.
For each entity $s$, the input is $\{\bm{x}_{s,t-m+1}, \bm{x}_{s,t-m+2},...,\bm{x}_{s,t}\}$, where $m$ is the size of time window.
We use a temporal self-attention layer to integrate the representations of entities in the temporal dimension, and the output embedding sequence for entity $s$ is defined as $\{\bm{z}_{s,t-m+1}, \bm{z}_{s,t-m+2},...,\bm{z}_{s,t}\}$.
We perform the scaled dot-product form of attention \cite{transformer} over the input embeddings at each time step, to generate the time-dependent embedding $\bm{z}_{s,t}$:
\begin{equation}
    e_{ij} = \frac{((\bm{X}\bm{W}_q)(\bm{X}\bm{W}_k)^T)_{ij}} {\sqrt{d}} + \bm{M}_{ij},
\end{equation}
\begin{equation}
    \beta_{ij} = \frac{{\rm exp}( e_{ij})} {\sum_{j'=0}^{m}{\rm exp}( e_{ij'})},
\end{equation}
\begin{equation}
    \bm{Z} = \bm{\beta}({\bm X}{\bm W}_v),
\end{equation}
where $\bm{X} \in {\rm R}^{m \times d}$ and  $\bm{Z} \in {\rm R}^{m \times d}$ are the input and output representations packed together across time.
$\bm{\beta} \in {\rm R}^{m \times m}$ is the attention weight matrix obtained by the multiplicative attention function.
And $\bm{W}_q \in {\rm R}^{d \times d}$, $\bm{W}_k \in {\rm R}^{d \times d}$ and $\bm{W}_v \in {\rm R}^{d \times d}$ are the linear projection matrices corresponding to the queries, keys, and values.
To ensure that future information is not exposed, we use the mask matrix $\bm{M} \in {\rm R}^{m \times m}$  as 
\begin{equation}
    \bm{M}_{ij} =\left\{
\begin{array}{rcl}
0,       &     & i \le j,\\
-\infty,       &     & otherwise.
\end{array}
	\right.
\end{equation}
In this way, when $i$ \textgreater $j$, the attention weight $\beta_{ij} \rightarrow 0 $, which enables that only representations in the past are assigned non-zero weights. 

As the SIP module is used to model the neighborhood interactions,
the TIP module focuses more on fine-grained changes, i.e., the events (relations) that occur between specific entity pairs across the time steps.
For each entity pair $(s,o)$, we aim to predict what events may happen. 
So we use the sequence of relations between $s$ and $o$, i.e., $\{r_{so}^0, r_{so}^1,..., r_{so}^n,...,r_{so}^{N}\}(1 \le n \le N)$, where $N$ is the number of the events that occur in the time window.
And relations are sorted by occurrence time. We use a gated recurrent unit (GRU) \cite{RNN} to get the relation transition representation for entity pair $(s,o)$ as follows:
\begin{equation}
\bm{z}_{so,n} = {\rm GRU}(\bm{x}_{so,n},\bm{z}_{so,n-1}),
\end{equation}
where $\bm{x}_{so,n} \in {\rm R}^d$ is the structural representation of relation $r_{so}^n$ at corresponding time.
Noting that an entity pair can have multiple events at the same time step, we simply follow the order in the raw data set.
And we take the output of the last hidden layer as the temporal relation representation $\bm{z}_{so,t}$.

\subsection{History Forward Module}
In this section, we introduce our scoring functions, which can evaluate an event from the structural, temporal and repetitive perspectives.
And we proposed a multi-step inference algorithm, which can generate new graphs sequentially for future time step to handle the extrapolation reasoning problem.

The temporal scoring function is used to predict what events may occur.
So we use the output from the TIP module as:
\begin{equation}
{I}_t(s,r,o,t) = {\rm softmax}(\bm{W}_t[\bm{z}_{s,t};\bm{z}_{so,t};\bm{z}_{o,t}])_r,
\label{temporal}
\end{equation}
where $\bm{W}_t \in {\rm R}^{p \times 3d}$ and $p$ is the number of relation types.
In this way, we can focus on the temporal evolution of events between specific entity pairs to predict the next event.

The structural scoring function evaluates events based on the structural representations, to consider the interactions between them at the same time step.
And borrowing scoring function from static KGE methods,
we use DistMult \cite{DistMult} to assign scores for the quadruples as follows:
\begin{equation}
{I}_s(s,r,o,t) = {\rm \sigma}(\langle\bm{x}_{s,t}, \bm{x}_{r,t}, \bm{x}_{o,t}\rangle),
\label{structural}
\end{equation}
where ${\rm \sigma}$ is the ${\rm sigmoid}$ function.
And $\langle\cdot\rangle$ denotes the tri-linear dot product.
Through the above two scoring functions, our method can consider the evolution patterns of events between entity pairs across time and neighborhood interactions at the same time step.

However, the existence of entities and events is time-sensitive, which may lead to less information available for some entities in the time window.
And based on the observation that many facts occur repeatedly along the history \cite{CyGNet},
we can generate new events from the historical events selectively to solve this problem.
In our method, we use historical vocabulary for each entity and relation.
For a query $(s,r,?,t)$, we compute the repetitive history score as follows:
\begin{equation}
{I}_h(s,r,o,t) = {\rm softmax}(\bm{W}_h[\bm{e}_{s,t};\bm{e}_{r,t}]+\bm{V}_{t}^{(s,r)})_o,
\label{vocabulary}
\end{equation}
\begin{equation}
\bm{V}_{t}^{(s,r)} = \bm{v}_1^{(s,r)}+\bm{v}_2^{(s,r)}+...+\bm{v}_{t-1}^{(s,r)},
\end{equation}
where $\bm{W}_h \in {\rm R}^{q \times 2d}$ and $q$ is the number of entities.
$\bm{e}_{s,t} \in {\rm R}^{d}$ and $\bm{e}_{r,t} \in {\rm R}^{d}$ are representations of entities and relations to model the preference of the history, which are independent of time and structure.
And $\bm{v}_{t^-}^{(s,r) }$ is an $q$-dimensional multi-hot indicator vector for subject $s$ and relation $r$,
and position $o$ is 1 represents that $(s,r,o,t^-)$ exists in graph  $\mathcal{G}^{(t^-)}$.

\begin{algorithm}[tb]
\caption{Reasoning algorithm of HIP network}
\label{alg:algorithm}
\textbf{Input}: Historical graph sequence $\{\mathcal{G}^{(1)},  \mathcal{G}^{(2)}, ...,  \mathcal{G}^{(t)}\}$,\\
Query set $\mathcal{S}_{query}$ with object entities missing at time $t + \Delta t$.\\ 
\textbf{Output}: The reasoning results for each query in descending order of scores. 
\begin{algorithmic}[1] 
\STATE $t'=t+1$.
\WHILE{$t' < t + \Delta t$}
\STATE Structural embeddings for entities and relations. $\triangleright$SIP
\STATE Temporal embeddings for entities. $\triangleright$TIP
\FOR {each $(s,r,?,t + \Delta t) \in \mathcal{S}_{query}$}
\STATE Generate candidate entity pair set $\mathcal{S}_{ep}^{t'}(s)$.
\STATE Generate candidate quadruple set $\mathcal{S}_{q}^{t'}(s)$. $\triangleright {\rm Eq.}$ \ref{temporal}
\STATE Pick top-$k$ quadruples in $\mathcal{S}_{q}^{t'}(s)$ and add them to $\mathcal{G}^{(t')}$. $\triangleright{\rm Eq.}$ \ref{structural} and \ref{vocabulary}
\ENDFOR
\STATE Add $\mathcal{G}^{(t')}$ to the graph sequence and update the historical vocabulary.
\ENDWHILE
\STATE Update structural embeddings.
\STATE Replace the missing part with all entities for each query and compute scores for them. $\triangleright{\rm Eq.}$ \ref{structural} and \ref{vocabulary}
\end{algorithmic}
\end{algorithm}

Now our goal is to handle the extrapolation reasoning problem with these scoring functions.
And the reasoning process during testing is shown in Algorithm 1 and Figure~\ref{fig:framework}.
At each time step, the first thing is to update representations using the SIP and TIP module (line 3-4).
But we compute the temporal embeddings for relations later.
Then the reasoning for queries can be divided into three main steps (line 5-10).

\emph{Step 1:}
When given a query $(s,r,?,t+\Delta t)$ and current time is $t'$,
we don't take all entities in $\mathcal{E}$. 
We generate the \textbf{candidate entity pair set} $\mathcal{S}_{ep}^{t'}(s)$ by searching for all entities that had events with $s$,
to reduce noise and computed space.

\emph{Step 2:}
Our method focuses on temporal information in this step.
We get the temporal embeddings for relations and use equation \ref{temporal} to find the most likely event for each entity pair in $\mathcal{S}_{ep}^{t'}(s)$.
We put the event with the highest score of each entity pair into \textbf{candidate quadruple} set $\mathcal{S}_{q}^{t'}(s)$.
This step enables our approach to incorporate temporal evolution, which can avoid predicting the same result at every step.

\emph{Step 3:}
We use the sum of equation \ref{structural} and \ref{vocabulary} to evaluate the plausibility of quadruples in $\mathcal{S}_{q}^{t'}(s)$,
which consider the preferences of current structure and history.
We choose the top-$k$ quadruples for each query, add them to the predicted graph $\mathcal{G}^{(t')}$, and then move on to the next time step $t'+1$.

To answer the queries,
we use the sum of equation \ref{structural} and \ref{vocabulary} to assign scores for each query with the missing part completed at target time step $t+ \Delta t$ (line 13), and rank them in descending order.

\subsection{Training Objective}

As the query $(s,r,?,t)$ for an object entity can be seen as a multi-class classification problem,
where each class corresponds to each object entity.
And if we need to predict what events will happen between entity pairs, we can also regard this as a multi-classification problem for relation types.
So the loss function in each time step can be defined as follows:
\begin{equation}
\mathcal{L} = \sum_{(s,r,o,t) \in \mathcal{G}^{(t)}}\sum_{*}{\rm -log}(I_*(s,r,o,t)).
\end{equation}
Here, $I_*(\cdot)$ are the scoring functions defined in the previous section, i.e., equation \ref{temporal}, \ref{structural} and \ref{vocabulary}.

\begin{table}
	\resizebox{\linewidth}{!}{
    \begin{tabular}{l|rr|rrrr}  
    \toprule
	Dataset & Entities & Relations & Training & Validation & Test & Time gap\\  
    \midrule
	YAGO & 10,623 & 10 & 161,540 & 19,523 & 20,026 & 1 year\\
	WIKI & 12,554 & 24 & 539,286 & 67,538 & 63,110 & 1 year\\
	ICEWS14 & 12,498 & 260 & 323,895 & - & 341,409 & 1 day\\
	ICEWS18 & 23,033 & 256 & 373,018 & 45,995 & 49,545 & 1 day\\
	GDELT & 7,691 & 240 & 1,734,399 & 238,765 & 305,241 & 15 mins\\
    \bottomrule
    \end{tabular}}
    \caption{Statistics of five datasets.}
\label{data}
\end{table}

\section{Experiments}

In this section, we evaluate our proposed framework, HIP network, on five public datasets.

\begin{table*}
	\resizebox{\textwidth}{!}{
    \begin{tabular}{l|rrrr|rrrr|rrrr|rrrr|rrrr|rrrr}  
    \toprule
			&\multicolumn{4}{c|}{YAGO}&\multicolumn{4}{c}{WIKI}&\multicolumn{4}{c}{ICEWS14}&\multicolumn{4}{c}{ICEWS18}&\multicolumn{4}{c}{GDELT}\\  
	Method & MRR & Hits@1 & Hits@3 & Hits@10 & MRR & Hits@1 & Hits@3 & Hits@10 & MRR & Hits@1 & Hits@3 & Hits@10 & MRR & Hits@1 & Hits@3 & Hits@10 & MRR & Hits@1 & Hits@3 & Hits@10\\  
    \midrule
	TransE & 48.97 & 46.23 & 62.45 & 66.05 & 46.68 & 36.19 & 49.71 & 51.71 & 18.65 & 1.12 & 31.34 & 47.07 & 17.56 & 2.48 & 26.95 & 43.87 & 16.05 & 0.00 & 26.10 & 42.29\\
	DistMult & 59.47 & 52.97 & 60.91 & 65.26 & 46.12 & 37.24 & 49.81 & 51.38 & 19.06 & 10.09 & 22.00 & 36.41 & 22.16 & 12.13 & 26.00 & 42.18 & 18.71 & 11.59 & 20.05 & 32.55\\
	ComplEx & 61.29 & 54.88 & 62.28 & 66.82 & 47.84 & 38.15 & 50.08 & 51.39 & 24.47 & 16.13 & 27.49 & 41.09 & 30.09 & 21.88 & 34.15 & 45.96 & 22.77 & 15.77 & 24.05 & 36.33\\
	ConvE & 62.32 & 56.19 & 63.97 & 65.60 & 47.57 & 38.76 & 50.10 & 50.53 & 40.73 & 33.20 & 43.92 & 54.35 & 36.67 & 28.51 & 39.80 & 50.69 & 35.99 & 27.05 & 39.32 & 49.44\\
	RotatE & 65.09 & 57.13 & 65.67 & 66.16 & 50.67 & 40.88 & 50.71 & 50.88 & 29.56 & 22.14 & 32.92 & 42.68 & 23.10 & 14.33 & 27.61 & 38.72 & 22.33 & 16.68 & 23.89 & 32.29\\
	RGCN & 41.30 & 32.56 & 44.44 & 52.68 & 37.57 & 28.15 & 39.66 & 41.90 & 26.31 & 18.23 & 30.43 & 45.34 & 23.19 & 16.36 & 25.34 & 36.48 & 23.31 & 17.24 & 24.96 & 34.36\\
	CompGCN & 41.42 & 32.63 & 44.59 & 52.81 & 37.64 & 28.33 & 39.87 & 42.03 & 26.46 & 18.38 & 30.64 & 45.61 & 23.31 & 16.52 & 25.37 & 36.61 & 23.46 & 16.65 & 25.54 & 34.58\\
	\midrule
	TTransE & 32.57 & 27.94 & 43.39 & 53.37 & 31.74 & 22.57 & 36.25 & 43.45 & 6.35 & 1.23 & 5.80 & 16.65 & 8.36 & 1.94 & 8.71 & 21.93 & 5.52 & 0.47 & 5.01 & 15.27\\
	HyTE & 23.16 & 12.85 & 45.74 & 51.94 & 43.02 & 34.29 & 45.12 & 49.49 & 11.48 & 5.64 & 13.04 & 22.51 & 7.31 & 3.10 & 7.50 & 14.95 & 6.37 & 0.00 & 6.72 & 18.63\\
	TA-DistMult & 61.72 & 52.98 & 63.32 & 65.19 & 48.09 & 38.71 & 49.51 & 51.70 & 20.78 & 13.43 & 22.80 & 35.26 & 28.53 & 20.30 & 31.57 & 44.96 & 29.35 & 22.11 & 31.56 & 41.39\\
	DySAT & 43.43 & 31.87 & 43.67 & 46.49 & 31.82 & 22.07 & 26.59 & 35.59 & 18.74 & 12.23 & 19.65 & 21.17 & 19.95 & 14.42 & 23.67 & 26.67 & 23.34 & 14.96 & 22.57 & 27.83\\
	TeMP & 62.25 & 55.39 & 64.63 & 66.12 & 49.61 & 46.96 & 50.24 & 51.81 & 43.13 & 35.67 & 45.79 & 56.12 & 40.48 & 33.97 & 42.63 & 52.38 & 37.56 & 29.82 & 40.15 & 48.60\\
	\midrule
	RE-NET & 65.16 & 63.29 & 65.63 & 68.08 & 51.97 & 48.01 & 52.07 & 53.91 & 45.71 & 38.42 & 49.06 & 59.12 & 42.93 & 36.19 & 45.47 & 55.80 & 40.12 & 32.43 & 43.40 & 53.80\\
	CyGNet & 66.58 & 64.26 & 67.98 & 70.16 & 52.60 & 50.48 & 53.26 & 55.82 & 49.89 & 43.15 & 53.68 & 61.18 & 47.83 & 42.02 & 50.71 & 57.72 & 51.06 & 44.66 & 54.74 & 61.32\\
	\midrule
	\textbf{HIP network} & \textbf{67.55} & \textbf{66.32} & \textbf{68.49} & \textbf{70.37} & \textbf{54.71} & \textbf{53.82} & \textbf{54.73} & \textbf{56.46} 
	& \textbf{50.57} & \textbf{45.73} & \textbf{54.28} & \textbf{61.65} 
	& \textbf{48.37} & \textbf{43.51} & \textbf{51.32} & \textbf{58.49} 
	& \textbf{52.76} & \textbf{46.35} & \textbf{55.31} & \textbf{61.87}\\
    \bottomrule
    \end{tabular}}
    \caption{Performance comparison on temporal link prediction.}
\label{main result}
\end{table*}

\subsection{Experimental Setup}

\paragraph{Datasets.}
We use five TKG datasets, namely ICEWS14 \cite{ICEWS14}, ICEWS18 \cite{ICEWS18}, GDELT \cite{GDELT}, WIKI \cite{WIKI} and YAGO \cite{YAGO}.
ICEWS14, ICEWS18 and GDELT are event-based TKGs which record events with timestamps.
WIKI and YAGO are subsets of Wikipedia history and YAGO3 respectively.
And we preprocess these datasets for extrapolation reasoning task as prior works \cite{RENET,CyGNet}: 
we split them into three subsets by timestamps except ICEWS14, i.e., train(80\%)/valid(10\%)/test(10\%).
Thus, (timestamps of train) $<$ (timestamps of valid) $<$ (timestamps of test).
More details about the five datasets can be found in Table~\ref{data}.

\paragraph{Evaluation metrics.}
We evaluate our method on the link prediction task which evaluates whether the ground-truth entity is ranked ahead of other entities. 
We report the results of mean reciprocal rank (MRR), hits at 1/3/10 (Hits@1/3/10) in our experiments.
And we remove corrupted entities as other baselines during evaluation which is called $filtered$ setting. 

\paragraph{Baselines.}
We mainly focus on comparing to the methods of static KGs and temporal graphs as prior works.
Static KG learning methods include TransE \cite{TransE}, DistMult \cite{DistMult}, ComplEx \cite{CompIEX}, ConvE \cite{ConvE}, RotatE \cite{RotatE}, RGCN-DistMult \cite{RGCN}, and CompGCN-DistMult \cite{CompGCN}.
And temporal reasoning methods include TTransE \cite{TTransE}, HyTE \cite{HyTE}, TA-DistMult \cite{TA-DistMult}, DySAT \cite{DySAT}, TeMP \cite{TeMP}, RE-NET \cite{RENET} and CyGNet \cite{CyGNet}.
Since RE-NET and CyGNet are the most relevant methods for extrapolation reasoning,
we will do more analysis on them.

\subsection{Results on TKGs}
In this task, we finetune the hyperparameters according to the MRR performance on each validation set.
Since ICEWS14 is not provided with a validation set, we use the same settings in ICEWS18.
And to be consistent with the baselines, most hyperparameters are the same on all datasets.
We use the ADAM optimizer with a learning rate of 0.001 to minimize the global loss.
The embedding dimension is 200. The dropout rate is 0.5. The batch size is 1024. And the time window size is set to 10.
Additionally, in our SIP module, we use multiplication as the composition operation and adopt 4 channels.
For the static KG methods, we simply remove all timestamps in datasets.

As Table~\ref{main result} shows, our HIP network outperforms all the baselines.
All static methods perform worse than our method since they do not consider temporal factors.
And TTransE, HyTE, TA-DistMult and DySAT don't even work as well as some static methods.
RE-NET, TeMP, CyGNet and our method outperform other methods by a large margin.
TeMP is not primarily proposed to solve the extrapolation reasoning problem,
but it's also trained along history,
which proves that modeling historical evolution patterns can better deal with this problem.
And compared to the best baseline CyGNet, 
our improvements on MRR and Hits@1 are more significant than Hits@10.
This may be because CyGNet generates predictions based on the historical distribution,
which can narrow the scope of predictions.
And some entities never appear in the training set, which also makes it difficult to improve Hits@10.
The improvements on Hits@1 and MRR fully demonstrate that
our method is able to utilize historical information more effectively, 
to predict exactly what is going to happen.

\begin{table}
	\resizebox{\linewidth}{!}{
    \begin{tabular}{l|rrrr|rrrr}  
    \toprule
			&\multicolumn{4}{c|}{WIKI}&\multicolumn{4}{c}{ICEWS14}\\  
	Metric & MRR & Hits@1 & Hits@3 & Hits@10 & MRR & Hits@1 & Hits@3 & Hits@10\\  
    \midrule
	SIP only & 48.25 & 39.17 & 50.36 & 52.11 & 31.81 & 26.17 & 33.65 & 48.59\\
	Vocabulary only & 52.89 & 50.62 & 53.39 & 55.96 & 49.14 & 44.23 & 52.27 & 58.92\\
	Vocabulary and SIP & 54.48 & 53.64 & 53.81 & 56.03 & 49.47 & 45.02 & 52.42 & 59.13\\
	\midrule
	\textbf{HIP network} & \textbf{54.71} & \textbf{53.82} & \textbf{54.73} & \textbf{56.46} 
	& \textbf{50.57} & \textbf{45.73} & \textbf{54.28} & \textbf{61.65}\\
    \bottomrule
    \end{tabular}}
    \caption{Effectiveness of each component on WIKI and ICEWS14.}
\label{ablation}
\end{table}

\subsection{Ablation Study}

We conduct ablation experiments on the WIKI and ICEWS14 datasets.
Firstly, we only use the SIP module and score the events through equation \ref{structural}.
Then we consider historical vocabulary, i.e., equation \ref{vocabulary}.
In the third setting, we combine historical vocabulary with structural representations to measure the plausibility of events.
Finally, we evaluate our method with all components according to Algorithm 1.

As shown in Table~\ref{ablation}, under the first setting, our method still performs better than most static KGE models, especially CompGCN.
This proves the usefulness of our SIP module which updates representations in a disentangled manner,
as changes often occur in the part of the neighborhood structure on TKGs.
And in the second setting, scoring with equation \ref{vocabulary},
our model can achieve better results,
which demonstrates the importance of selectively incorporating historical information for each query.
When we combine the two settings together, the result on all metrics still improves.
As we consider the evolution patterns about events between entity pairs,
HIP network achieves the best results by using the multi-step reasoning process.
In particular, the improvements in the purely event-based dataset (ICEWS14) are more significant,
this may be because events in the ICEWS14 have more evolution patterns (the time gap in ICEWS14 is only 24 hours),
which proves the importance of considering the temporal evolution of events.

\subsection{Sensitivity Analysis}
In this section, we investigate the influence of hyperparameters on YAGO.

\paragraph{Number of channels in the structural aggregator.}

\begin{figure}
    \centering
    \subfigure[Number of channels]{
        \centering
    \label{k}
    \includegraphics[width=1.60in]{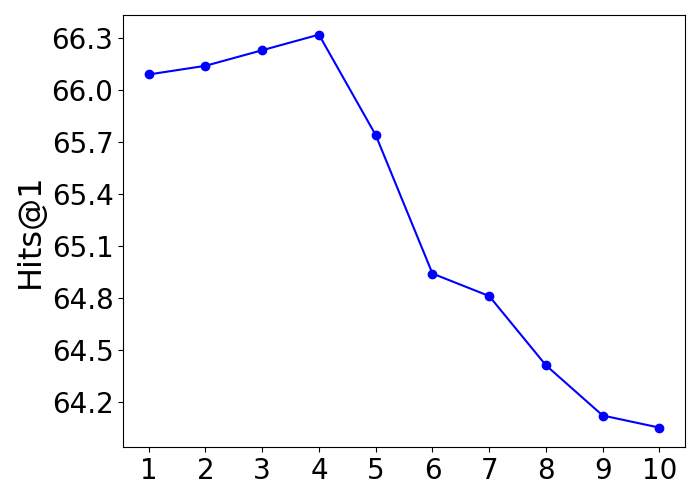}
    }    
    \subfigure[Length of history]{
        \centering
    \label{m}
    \includegraphics[width=1.60in]{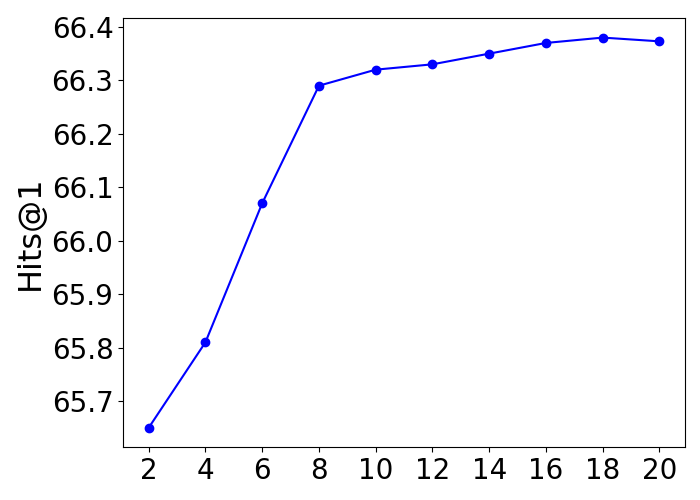}
    }
    \caption{Parameter sensitivity on HIP network.}
    \label{fig:sensitivity} 
\end{figure}

To update the related parts of embeddings in the aggregated process,
we project the embeddings and the results of composition operation onto $K$ subspaces.
We consider the value of $K$ from 1 to 10.
As Figure~\ref{fig:sensitivity}(a) shows, our model achieves the best result when $K$ is set to 4.
And our model is sensitive to $K$ especially when $K$ is larger than 4.
Since the weight of each subspace is calculated in a similar way to the multi-head attention mechanism,
large $K$ results in a smaller subspace dimension.
So we need to choose $K$ more carefully according to the embedding dimension.

\paragraph{Length of history in the temporal part.}
The input of the TIP module is the information in the time window.
In detail, we take the sequence of events up to $m$ between each entity pair.
Note that there may be multiple events between each entity pair at the same time step, we take the $m$ events that have recently occurred in the order of the original dataset.
And we simply remove the above events that do not occur in the time window.
So $m$ is the limit on the number and scope of events.
And in the temporal self-attention layer, we only consider the embeddings within the time window.
Figure~\ref{fig:sensitivity}(b) shows that the longer the history is considered, the higher the Hits@1.
When the history length is around 8, Hits@1 starts to be relatively stable.
And a large time window may result in a long sequence of events in RNN.
Therefore, to simplify the computation and save time, we don't need to set $m$ too large.

\section{Conclusions}

In this paper, we propose HIP network to solve the extrapolation reasoning problem on TKGs.
We pass historical information from the temporal, structural and repetitive perspectives to make future predictions.
And the proposed HIP network not only considers the important role of relations in the information passing process,
but also evaluates the plausibility of an event from above perspectives, which can effectively model the evolution patterns, neighborhood interactions, and historical repetition.
Experimental results show that HIP network achieves improvements over state-of-the-art baselines.

\section*{Acknowledgments}

This work was supported by the National Key R\&D Program with No.2016QY03D0503,2016YFB081304, and Strategic Priority Research Program of Chinese Academy of Sciences, Grant No.XDC02040400.

\clearpage
\bibliographystyle{named}
\bibliography{ijcai21}

\end{document}